\documentclass[11pt]{article}

\usepackage{latexsym,multirow}
\usepackage{amssymb,amsmath, bm}
\usepackage{cancel}
\usepackage{graphicx}
\usepackage{booktabs}

\usepackage{enumerate}

\usepackage{tikz}
\usetikzlibrary{bayesnet}

\usepackage{natbib}
\usepackage[pdftex, bookmarksopen=true, bookmarksnumbered=true,
pdfstartview=FitH, breaklinks=true, urlbordercolor={0 1 0}, citebordercolor={0 0 1}]{hyperref}
\usepackage{colortbl}
\usepackage{subcaption}

\usepackage{dcolumn}
\newcolumntype{.}{D{.}{.}{-1}}
\newcolumntype{d}[1]{D{.}{.}{#1}}
\usepackage{theorem}
\theoremstyle{plain}
\theoremheaderfont{\scshape}
\newtheorem{assumption}{Assumption}

\newcommand{\ind}{\mbox{$\perp\!\!\!\perp$}}

\newcommand{\argmax}{\operatornamewithlimits{argmax}}

\usepackage{rotating}

\usepackage[compact]{titlesec}

\allowdisplaybreaks

\newcommand\spacingset[1]{\renewcommand{\baselinestretch}%
  {#1}\small\normalsize}

\let\S\relax 
\DeclareRobustCommand{\S}{%
  \ifmmode
    \mathsection
  \else
    \textsection~%
  \fi
}
\usepackage{versions}
\excludeversion{synth}
\includeversion{real}
\graphicspath{{figs/}}

\newcommand{\blind}{0}

\newcommand{\Beta}{\text{Beta}}

\newcommand{\Dir}{\text{Dirichlet}}

\newcommand{\wtphi}{\widetilde{\bm{\phi}}}

\newcommand{\E}{\mathbb{E}}

\newcommand{\bbeta}{\bm{\beta}}

\newcommand{\bphi}{\bm{\phi}}
\newcommand{\btheta}{\bm{\theta}}
\newcommand{\bgamma}{\bm{\gamma}}

\begin{document} 

\newcommand{\tit}{Addressing Census data problems in race imputation
  via fully Bayesian Improved Surname Geocoding\\ and name supplements}


\spacingset{1.25}

\if0\blind

{\title{{\bf\tit}\thanks{We thank Bruce Willsie of L2, Inc. for his
      generosity in letting us make publicly available additional data
      on names and race.}}

  \author{Kosuke Imai\thanks{Professor, Department of Government and Department of
      Statistics, Harvard University.  1737 Cambridge Street,
      Institute for Quantitative Social Science, Cambridge MA 02138.
      Email: \href{mailto:imai@harvard.edu}{imai@harvard.edu} URL:
      \href{https://imai.fas.harvard.edu}{https://imai.fas.harvard.edu}}
    \and Santiago Olivella\thanks{Associate Professor, Department of 
      Political Science, University of North Carolina at Chapel Hill.}
    \and Evan Rosenman\thanks{Postdoc, Harvard Data Science Initiative, Harvard University.}}

  \date{\today
}

\maketitle

}\fi

\if1\blind
\title{\bf \tit}

\maketitle
\fi

\pdfbookmark[1]{Title Page}{Title Page}

\thispagestyle{empty}
\setcounter{page}{0}
         
\begin{abstract}
  Prediction of individual's race and ethnicity plays an important
  role in social science and public health research.  Examples include
  studies of racial disparity in health and voting.  Recently,
  Bayesian Improved Surname Geocoding (BISG), which uses Bayes' rule
  to combine information from Census surname files with the geocoding
  of an individual's residence, has emerged as a leading methodology
  for this prediction task.  Unfortunately, BISG suffers from two
  Census data problems that contribute to unsatisfactory predictive
  performance for minorities.  First, the decennial Census often
  contains zero counts for minority racial groups in the Census blocks
  where some members of those groups reside.  Second, because the
  Census surname files only include frequent names, many surnames --
  especially those of minorities -- are missing from the list. To
  address the zero counts problem, we introduce a fully Bayesian
  Improved Surname Geocoding (fBISG) methodology that accounts for
  potential measurement error in Census counts by extending the
  na\"ive Bayesian inference of the BISG methodology to full posterior
  inference.  To address the missing surname problem, we supplement
  the Census surname data with additional data on last, first, and
  middle names taken from the voter files of six Southern states where
  self-reported race is available.  Our empirical validation shows
  that the fBISG methodology and name supplements significantly
  improve the accuracy of race imputation across all racial groups,
  and especially for Asians.  The proposed methodology, together with
  additional name data, is available via the open-source software
  package \textsc{wru}.


\end{abstract}


\clearpage 

\section{Introduction}

Social scientists and public health researchers often must predict
individual race and ethnicity when assessing disparities in policy and
health outcomes.  The Bayesian Improved Surname Geocoding (BISG),
which uses Bayes' rule to combine information from the Census surname
list with the geocoding of individual residence, has emerged as a
leading methodology for this prediction task
\citep{elli:etal:08,elli:etal:09,fisc:frem:06,imai2016improving}.
Recent applications of the BISG methodology include studies on racial
disparity in police violence \citep{edwa:lee:espo:19}, eviction
\citep{hepb:loui:desm:20}, suicide \citep{stud:etal:20}, and turnout
\citep{frag:18}.

In this paper, we address two Census data problems that hinder
accurate prediction of individual race and ethnicity when using the
BISG.  First, the decennial Census often contains zero counts for
minority groups in the Census blocks where some members of those
groups reside.  This may happen for several reasons.  Some individuals
may have moved after the decennial Census.  There may also be
under-counts.  Another possibility is that the Census may inject
measurement error for privacy protection \citep[see][and references
therein]{kenny2021use}.  

Second, the decennial Census surname files only include the racial
composition of surnames that occur 100 or more times in the
population.  According to the Census Bureau, these names account for
about 90 percent of people with surnames recorded in the 2010 Census
\citep{come:16}.  This means that no racial breakdown statistic is
available for the remaining 10 percent.  This lack of information may
disproportionately affect minority groups if their surnames are less
frequently occurring than those of the majority group.

Using data from six Southern states in which individual race of voters
is available for validation, we show how these problems can result in
a deterioration of predictive quality for the standard BISG approach
(Section~\ref{sec:problems}).  To address these problems, we introduce
a fully Bayesian generalization of the BISG approach and extend the
coverage of available name-race tables (Section~\ref{sec:methods}).
  
Our empirical validation study demonstrates that these proposed
solutions yield substantial improvements in predictive accuracy ---
particularly among racial minorities
(Section~\ref{sec:empirics}). Specifically, a model that incorporates
all our proposed improvements increases classification accuracy by an
average of about 14\% among all five major racial groups
(\emph{vis-\`a-vis} the standard BISG), with improvements as high as
26\% among Asian voters. Moreover, these gains in predictive accuracy
do not come at the expense of the calibration of predicted
probabilities across racial groups, which is particularly high for
predictions made by the standard BISG methodology for White and Black
voters. Finally, we conclude with a brief discussion about the
applicability of our proposed modeling approach to various domains.


\section{The Census Data Problems in Race Imputation}
\label{sec:problems}

In this section, we first briefly review the standard BISG
methodology.  We then describe the census data problems and quantify
the degree to which they negatively affect the predictive performance
of BISG.

\subsection{Bayesian Improved Surname Geocoding (BISG): A Review}

The goal of BISG is to predict the race of individual $i$, defined as
$R_i \in \mathcal{R}$ where $|\mathcal{R}| = J$ is the total number of
(mutually exclusive) racial categories. In this manuscript, we will
have $J = 5$, with the categories $\mathcal{R} = \{$``White,"
``Black," ``Hispanic," ``Asian," ``Other"$\}$.

Suppose we observe the individual's surname
$S_i \in \mathcal{S}=\{1,2,\ldots,K\}$ and geolocation
$G_i \in \mathcal{G}=\{1,2,\ldots,L\}$ where the latter is typically
recorded as a Census geographical unit (e.g., Census block), in which
his or her residence is located.  The BISG methodology is an
application of na\"ive Bayes prediction, where the key assumption is
given by the following conditional independence relation between
geolocation and surname, given race.
\begin{assumption}\label{asm:sur-geo} {\sc (Independence between
    Surname and Geolocation within Racial Group)}\spacingset{1}
$$G_i \ \ind \ S_i \mid R_i$$
\end{assumption}
Under Assumption~\ref{asm:sur-geo}, the BISG
prediction of an individual's race is given by,
\begin{align}
  \begin{split}
    P(R_i \mid S_i, G_i) &\ \propto \ P(S_i \mid R_i, G_i) P(R_i \mid
    G_i) \ = \ P(S_i \mid R_i) P(R_i \mid G_i).
\label{eq:BISG} 
 \end{split}
\end{align}
One may also use the following equivalent formula obtained via another
application of Bayes' rule: 
\[ P(R_i \mid S_i, G_i) \propto P(R_i \mid S_i) P(G_i \mid R_i) \]
In practice, the decennial Census surname files are used to compute
$P(S_i \mid R_i)$, whereas for $P(R_i \mid G_i)$ it is common to use
the Census Bureau's cross-tabulations of racial category by geographic
location (e.g. Census blocks).

Although we do not address the appropriateness of
Assumption~\ref{asm:sur-geo} in this paper, it is important to
acknowledge its limitation.  The assumption is violated if, for
example, among Asian Americans, various ethnic groups (Chinese,
Indians, Japanese, Korean, Vietnamese, etc.) have distinct surnames and
tend to live in different areas.  A similar problem might also arise
among Hispanic Americans.  The surname ``Santos,'' for instance, may
be common among Hispanics in some areas, but it may also be a common
last name among Brazilian Americans (who are classified as
non-Hispanic whites according to the Census) in other areas.

We now turn to two Census data problems that negatively affect
the predictive performance of BISG.

\subsection{Consequences of Zero Census Counts}

The decennial census is intended to provide a full accounting of where
each resident of the United States lives as of April 1 on the census
year. Reported census distributions are considered reasonably reliable
as of this date, though still imperfect \citep[see
e.g.][]{kenny2021use}. Over time, however, this accuracy degrades even
further, as individuals move within the nation's borders at
significant rates \citep{basso2020internal}. At fine levels of
resolution, such as Census blocks, this means that the racial
distributions may not fully capture the diversity of residents within
a short time after the census is conducted. Among rapidly growing
minority groups, such as Asian Americans and Hispanic Americans,
errors may be particularly large.

Prior studies have shown that the use of the Census block level data,
rather than the data at a higher level of geographical aggregation,
tend to yield more accurate BISG prediction of individual race and
ethnicity \citep[e.g.,][]{imai2016improving}.  This, however, can
result in a greater chance of measurement error.  In particular, when
Census counts are used to obtain the prior distribution
$P(R_i \mid G_i)$ at the Census block level, some blocks may record
zero individuals of certain ethnic and racial categories.  In such
cases, $P(R_i \mid G_i)$ would be set to zero, making the posterior
probability of belonging to these groups automatically zero for all
individuals who reside in these blocks.  For example, someone with the
last name ``Guti\'errez'' --- a distinctively Hispanic last name ---
living in a neighborhood where the Census failed to count anyone of
Hispanic descent would have a zero posterior probability of being
classified as such according to the standard BISG methodology.

Thus, the predictive accuracy of the BISG methodology can suffer
dramatically when probabilities are zeroed-out \emph{a priori}.  To
quantify this error, we consider the voter files of six Southern
states --- Alabama, Florida, Georgia, Louisiana, North Carolina, and
South Carolina --- sourced between October 2020 and February 2021
(prior to the release of 2020 census data). The voter files were
provided by L2, Inc., a leading national non-partisan firm and the
oldest organization in the United States that supplies voter data and
related technology to candidates, political parties, pollsters, and
consultants for use in campaigns.  These files tally all registered
voters (approximately 37.8 million voters) in the state as of the
production date, geocoding the Census blocks of their home addresses.
About 91\% of these voters provided self-reported race data.

\begin{table}[t!] \centering \spacingset{1}
\begin{tabular}{lccccc} \textbf{Census Tally} &
\multicolumn{1}{c}{\textbf{White}} &
\multicolumn{1}{c}{\textbf{Black}} &
\multicolumn{1}{c}{\textbf{Hispanic}} &
\multicolumn{1}{c}{\textbf{Asian}} &
\multicolumn{1}{c}{\textbf{Other}} \\ \toprule \textbf{Zero counts} & 0.12
(1\%) & 0.32 (4\%) & 0.16 (5\%) & 0.13 (20\%) & 0.22 (30\%) \\
\textbf{Non-zero counts} & 22.13 (99\%) & 7.55 (96\%) &
2.84 (95\%) & 0.51 (80\%) & 0.51 (70\%) \\ \bottomrule
\end{tabular}
\caption{\label{tab:zeroes} Count of individuals of each race in
  millions, for whom the 2010 Census states that there are zero (top
  row) or more than zero (bottom row) members of that racial group
  living within the individual's home census block. Data is sourced
  from voter files from AL, FL, GA, LA, NC, and SC and racial data is
  self-reported on the file.} 
\end{table}

Table~\ref{tab:zeroes} shows the counts of voters (in millions) by
self-reported race, further divided by whether the 2010 Census
indicates that exactly zero members of the individual's racial group
live within their home block. Due to internal mobility and other forms
of measurement error, just shy of one million voters (2.8\%) live in a
Census block where the 2010 Census tallies indicate that no members of
the individual's racial group reside.  Notably, these errors are not
shared evenly across races. While fewer than 1\% of White voters live
in a Census block in which the Census data says no White individuals
reside, a full fifth of Asian voters live in Census blocks in which
the 2010 Census says there are no Asian residents. Even these
aggregates mask substantial heterogeneity by state. In South Carolina,
for example, 19\% of Hispanic voters and 31\% of Asian voters reside
in zero-Hispanic and zero-Asian blocks, according to the 2010 Census.

This mismatch presents a significant challenge for the BISG
methodology.  A na\"ive application of BISG would yield a prediction
of 0\% for the true racial group of all individuals in the first row
of Table~\ref{tab:zeroes} --- comprising a relatively large proportion
of all minority voters in the South --- simply as a mechanical result
the Census reporting no members of these racial groups living in the
corresponding geographies.

\begin{table}[t] \centering \spacingset{1}
\begin{tabular}{lccccc} \textbf{Census Tally} & \textbf{White} &
\textbf{Black} & \textbf{Hispanic} & \textbf{Asian} & \textbf{Other}
\\ \toprule \textbf{Zero counts} & 50.0\% & 50.0\% & 50.0\% & 50.0\% & 50.0\%
\\ \textbf{Non-zero counts} & 89.2\% & 92.4\% & 94.9\% &
91.4\% & 58.9\% \\ \textbf{Overall} & 89.8\% & 91.7\% & 91.9\% &
82.2\% & 59.0\% \\ \bottomrule
\end{tabular}
\caption{\label{tab:zeroesRocAUC} Area under the receiver operating
characteristic curve (AUROC) for BISG predictions of individual race,
using Census blocks to set racial prior distributions. Overall AUROC
values (third row) are lower for all non-white racial groups than are
AUROC values for individuals living in blocks for which the racial
prior for those groups is nonzero.}
\end{table}

The impact on BISG prediction is summarized in
Table~\ref{tab:zeroesRocAUC}, where we split out the data as in
Table~\ref{tab:zeroes} and compute the area under the receiver
operating characteristic (AUROC) curve for the BISG predictions on
each subgroup.  The AUROC measures the probability that a randomly
chosen member of each racial group will have a higher predicted
probability of belonging to that racial group than a randomly chosen
non-member. Accordingly, higher values of the AUCROC indicate better
classification accuracy.  The first row of entries are all equal to
50.0\% --- a direct consequence of the fact that the true positive
rates must be zero for each racial group in any block for which the
Census prior is zero. In Census blocks for which the prior on the
racial group is nonzero, the AUROC is above 90\% for Black, Hispanic,
and Asian voters, indicating good predictive accuracy. As a result,
aggregate performance across \emph{all} Census blocks (the third row)
tends to be poorer for most racial groups (with the exception of White
and Other voters), owing to the poor prediction on Census blocks for
which the prior is erroneously set to zero.

We can also compute misclassification rates for each racial group, by
assigning each individual to the the maximum a posteriori class and
comparing against their true, self-reported race. These results can be
found in Table~\ref{tab:censusBlockPreds} in the Appendix, where we
report both overall error and false positive and false negative rates
by racial group.  All individuals living in Census blocks for whom the
prior equals zero for their true race are misclassified, driving the
overall error rate up from 14.5\% to 16.9\%.

These results suggest that individual race prediction can be improved
by addressing the possibility that block-level racial priors may be
inaccurate or out of date, especially if they are equal to zero.




\subsection{Consequences of Missing Race-Name Data}

A second plausible source of error arises from the use of surname
data. In most applications of the BISG methodology, surname racial
distributions are drawn from the Census Bureau’s surname list.  The
2010 Census surname list, for example, provides the racial
distribution of surnames appearing at least 100 times, which amounts
to a total of about 160,000 names. These data are supplemented with
the Census's Spanish surname list, a list of about 12,000 common
Hispanic surnames, approximately half of which are not in the Census
surname list.

\begin{table}[t!] \centering \spacingset{1}
\begin{tabular}{lccccc} \textbf{Name Match?}  &
\multicolumn{1}{c}{\textbf{White}} &
\multicolumn{1}{c}{\textbf{Black}} &
\multicolumn{1}{c}{\textbf{Hispanic}} &
\multicolumn{1}{c}{\textbf{Asian}} &
\multicolumn{1}{c}{\textbf{Other}} \\ \toprule \textbf{No} & 1.47
(7\%) & 0.27 (3\%) & 0.13 (4\%) & 0.09 (14\%) & 0.08 (10\%) \\
\textbf{Yes} & 20.79 (93\%) & 7.61 (97\%) & 2.88 (96\%) & 0.55 (86\%)
& 0.66 (90\%) \\ \bottomrule
\end{tabular}
\caption{\label{tab:zeroNames} Count of individuals (in millions) of
  each race for whom the individual's surname cannot be matched to a
  name in the Census surname dictionary or Hispanic surname file. Data
  is sourced from voter files from AL, FL, GA, LA, NC, and SC and
  racial data is self-reported on the file.}
\end{table}

While these data are quite broad, they do not account for the 
possibility of rare surnames. In our sample of Southern states as of 
late 2020 and early 2021, we find that about 2 million voters (5.9\%) 
have surnames that cannot be matched to the census name dictionary, 
even after the data are cleaned and stripped of punctuation to improve 
the chance of a match. The distribution of this mismatch across racial 
groups is given in Table~\ref{tab:zeroNames}. Although Asian voters 
are particularly unlikely to have their surnames matched (14\%), the 
same is true for a significant portion of White voters (7\%). 

\begin{table}[t!] \centering \spacingset{1}
\begin{tabular}{lccccc} \textbf{Name Match?}  & \textbf{White} &
\textbf{Black} & \textbf{Hispanic} & \textbf{Asian} & \textbf{Other}
\\ \toprule \textbf{No} & 79.4\% & 85.5\% & 78.1\% & 71.3\% & 55.9\%
\\ \textbf{Yes} & 90.3\% & 91.8\% & 92.2\% & 82.3\% & 59.1\% \\
\textbf{Overall} & 89.8\% & 91.7\% & 91.9\% & 82.2\% & 59.0\% \\
\bottomrule
\end{tabular}
\caption{\label{tab:zeroNameRocAUC} Area under the receiver operating
  characteristic curve (AUROC) for BISG predictions of individual
  race, using Census blocks to set racial prior
  distributions. Predictive performance among individuals whose last
  names cannot be matched to the \texttt{WRU} name dictionary (first
  row) is significantly worse than among individuals for whom a match
  is found (second row). }
\end{table}

In the absence of a surname match, the default behavior of a common
implementation of BISG (viz. the software package \texttt{WRU}
\citep{wru}) is to use the approximate 2010 national race
proportions as an estimate for $P(R_i \mid S_i)$. This approximation
yields a degradation in predictive performance among these records, as
seen in Table~\ref{tab:zeroNameRocAUC}.  The AUROC is significantly
lower among individuals without a name match than among those whose
surnames are found in the dictionary, and the discrepancy is more
than ten percentage points for White, Hispanic, and Asian voters.

As in the prior section, we compute misclassification rates for each
racial group.  These results can be found in
Table~\ref{tab:missingNamePreds} in the Appendix.  The results are
somewhat less dramatic in this case, because the default behavior in
the absence of a name match does not automatically yield a
misclassification.  Nonetheless, we can see that misclassifications
occur for less than one sixth of individuals whose names are matched,
but nearly a quarter of individuals whose names are unmatched,
increasing the overall error rate.


Once again, a data limitation yields a reduction in the predictive
performance of the BISG methodology. Accordingly, better name coverage
would improve the quality of our predictions. In what follows, we lay
out our proposed solutions to these common data quality issues, and
show how correcting for them can substantially improve the prediction
accuracy of BISG. In addition, we further extend the common BISG
approach to incorporate first and middle names --- information that is
typically readily available from voter files, and which can further
improve BISG's accuracy.

\section{The Proposed Solutions}
\label{sec:methods}

In this section, we propose solutions to the measurement error
problems described above.  We begin by introducing a measurement error
model designed to address potential for error in census tallies.  Our
model generalizes the na\"ive Bayes BISG methodology to a fully
Bayesian model.  We complete our approach by discussing our name
augmentation strategy, designed to correct for lack of coverage in
commonly used name-by-race dictionaries.

\subsection{Accounting for the Measurement Error in Census Counts}

We use a fully Bayesian modeling strategy to account for potential
measurement error that arises when quantifying the racial distribution
within each geography.
We begin by modeling the observed Census counts as a draw from a
Multinomial distribution with the true, but unknown, race proportions
in geolocation $g$, denoted by
$\boldsymbol{\zeta}_g = (\zeta_{1g},\zeta_{2g},\ldots,\zeta_{Jg})$,
\begin{equation}
  \bm{N}_{g} \ \stackrel{\text{indep.}}{\sim} \ \text{Multinom}(N_g, \boldsymbol{\zeta}_{g})
\end{equation}
where $\bm{N}_g=(N_{1g},N_{2g},\ldots,N_{Jg})$ is the $J$-dimensional
vector of Census counts for individuals who belong to different racial
groups and live in geolocation $g$, and
$N_g = \sum_{r \in \mathcal{R}} N_{rg}$ is the observed total Census
population count in geolocation $g$.

Next, we place the following conjugate prior distribution over the
unknown race distribution for the geolocation $g$,
\begin{equation}
  \boldsymbol{\zeta}_{g} \ \stackrel{\text{indep.}}{\sim} \ \text{Dirichlet}(\bm{\alpha})
\end{equation}
where $\bm{\alpha}=(\alpha_{1},\alpha_{2},\ldots,\alpha_{J})$ is the
$J$-dimensional vector of prior hyperparameters. In our
implementation, we define a uniform prior distribution with
$\bm{\alpha}=\mathbf{1}$.  This provides enough smoothing over
observed zero counts, without preferring a particular ethnic and
racial group over another.

We call this measurement error model the fully-Bayesian BISG (fBISG).
Letting $P(S_i \mid R_i=r)=\boldsymbol{\pi}_{r}$, the full posterior
distribution of the fBISG is given by,
\begin{align}
  \begin{split}
 & P(\{R_i\}_{i=1}^n ,
\{\bm{\zeta}_g\}_{g\in\mathcal{G}} \mid \bm{S}, \bm{G},
                    \{\bm{\pi}_r\}_{r\in\mathcal{R}},\bm{\alpha})  \\
                 & \propto 
\prod_{i=1}^n \prod_{r \in \mathcal{R}} \prod_{g\in\mathcal{G}}\left\{
\left(\prod_{s \in \mathcal{S}} \pi_{sr}^{\mathbf{1}\{S_i =
s\}}\right) \zeta_{rg}^{\mathbf{1}\{G_i=g\}}\right\}^{\mathbf{1}\{R_i
= r\}} \times \prod_{r \in \mathcal{R}} \prod_{g \in
                             \mathcal{G}}\zeta_{rg}^{N_{rg}+\alpha_r-1} \\
                 & =  \prod_{s
\in \mathcal{S}} \prod_{r \in \mathcal{R}} \pi_{sr}^{m_{sr}} \times
\prod_{r \in \mathcal{R}} \prod_{g\in\mathcal{G}} \zeta_{rg}^{n_{rg} +
  N_{rg} + \alpha_{r} - 1}
\end{split}
\end{align}
where $n_{rg}=\sum_{i=1}^n\mathbf{1}\{R_i = r, G_i =
g\}$ is the number of individuals on the voter file who belong to race
$r$ and live in geographical unit $g$, and $m_{sr}=\sum_{i=1}^n
\mathbf{1}\{S_i = s, R_i = r\}$ is the number of individuals in the
voter file who belong to race $r$ and have surname $s$.

To simplify computation, we integrate out $\boldsymbol{\zeta}_g$,
obtaining the following marginalized posterior distribution,
\begin{align}
  \begin{split} P(\{R_i\}_{i=1}^n\mid
\{\bm{\pi}_r\}_{r\in\mathcal{R}}, \bm{S}, \bm{G}, \bm{\alpha}) & \
\propto \ \prod_{r\in\mathcal{R}} \left\{ \prod_{s \in \mathcal{S}}
\pi_{sr}^{m_{sr}} \prod_{g\in\mathcal{G}} \Gamma(n_{rg} + N_{rg} +
\alpha_{r}) \right\}.  \label{eq:post}
  \end{split}
\end{align}

To sample from this joint posterior, we construct a Gibbs sampler.
Using the fact that $\Gamma(x+y)=x^y\Gamma(x)$ for $y \in \{0, 1\}$,
we can derive the following conditional posterior distribution for
$R_i$ given the race of the other individuals,
\begin{equation} \Pr(R_i = r \mid \bm{R}_{-i}, S_i = s, G_i = g,
\bm{G}_{-i}, \boldsymbol{\alpha}) \ \propto \
\pi_{sr}(n_{rg}^{-i}+N_{rg}+\alpha_{r})
  \label{eq:gibbs}
\end{equation} where $n_{rg}^{-i}=\sum_{i^\prime \ne i}
\mathbf{1}\{R_{i^\prime} = r,G_{i^\prime} = g\}$ is the only parameter
that needs to be updated throughout the sampling process.
After the corresponding Markov chain has converged to its stationary
distribution, the posterior prediction for $R_i$ can now be based on
the posterior approximated by iteratively sampling from the full set
of conditional distributions in Equation~\eqref{eq:gibbs}.

The comparison of Equation~\eqref{eq:gibbs} with
Equation~\eqref{eq:BISG} shows how the fBISG addresses the problems
caused by zero census counts.  Notice that the race-geolocation
probability $\Pr(R_i \mid G_i)$ in the BISG prediction formula, which
is given by $N_{rg}/\sum_{r^\prime g}N_{r^\prime g}$, is replaced with
the ratio
$(n_{rg}^{-i} + N_{rg} + \alpha_r)/\sum_{r^\prime}(n_{r^\prime g}^{-i}
+ N_{r^\prime g} + \alpha_{r^\prime})$ in the fBISG formula.  Thus, in
the BISG methodology, if $N_{rg}=0$, the posterior prediction for this
racial group $r$ in the geolocation $g$ is zero.  In contrast, the
fBISG methodology gives non-zero probability of belonging to the
racial group with zero Census counts by adding a prior and partially
pooling other individuals who live in the same geolocation.

\subsection{Increasing Surname Coverage, and Incorporating First and Middle Names}
\label{subsec:names}

In addition to the surname, we may also have first and middle names of
each individual whose racial group we wish to predict. Let
$F_i \in \mathcal{F}=\{1,2,\ldots,K_F\}$ and
$M_i \in \mathcal{M}=\{1,2,\ldots,K_M\}$ denote the first and middle
names of individual $i$, respectively.  Using the same voter file data
from L2, Inc., we construct the racial composition of each first name
and that of each middle name.  This allows us to further approximate
the joint distributions $P(R_i, F_i)$ and $P(R_i, M_i)$.

\citet{voic:18} shows that incorporating the first name can improve
the performance of the BISG.  The author makes the assumption, similar
to Assumption~\ref{asm:sur-geo}, that the first name is independent of
geolocation conditional on race.  In addition, it is assumed that the
first name is independent of the surname given race.  If we make the
same assumption about middle names, the prediction formula becomes,
\begin{equation*} P(R_i \mid F_i, M_i, S_i, G_i) \ \propto \ P(F_i\mid
R_i) P(M_i \mid R_i) P( S_i \mid R_i) P(R_i \mid
G_i). \label{eq:BISG-first-middle}
\end{equation*} Combining this information with our fully-Bayesian
model for smoothing over zero census counts results in the following
updated full conditional distribution over individual $i$'s race:
\begin{equation} \Pr(R_i = r \mid \bm{R}_{-i}, F_i = f, M_i = m, S_i =
s, G_i = g, \bm{G}_{-i}, \boldsymbol{\alpha}) \ \propto \
\pi^{\mathcal{F}}_{fr}\pi^{\mathcal{M}}_{mr}\pi^{\mathcal{S}}_{sr}(n_{rg}^{-i}+N_{rg}+\alpha_{r})
  \label{eq:gibbs_sfm}
\end{equation} where $\pi^{\mathcal{F}}_{fr}= P(F_i=f\mid R_i=r)$, and
similarly with $\pi^{\mathcal{M}}_{mr}$ and $\pi^{\mathcal{S}}_{sr}$.

We next demonstrate the empirical benefit of incorporating voter file
data surname racial distributions, as well as those of first and
middle names. Using the same set of voter files from L2, Inc., we
consider matching individuals to name dictionaries under several
schemes. First, we consider surnames exclusively, and compute the
proportion of individuals from each racial group who do not have a
surname matched to the Census dictionaries (as in
Table~\ref{tab:zeroNames}). Next, we compute the proportion of
individuals of each race who do not have a surname matched to the
Census dictionaries, augmented with data from the L2 voter files
themselves. Third, we compute the proportion of individuals of each
race who do not have a surname matched to the augmented surname
dictionary \emph{or} a first name matched to the separate first name
dictionary compiled from the L2 data. Lastly, we compute the
proportion of individuals of each race who do not have a name matched
to any of the augmented surname dictionary or to first and middle name
dictionaries compiled from the L2 data.

\begin{figure}[t] \centering \spacingset{1}
  \includegraphics[width=\textwidth]{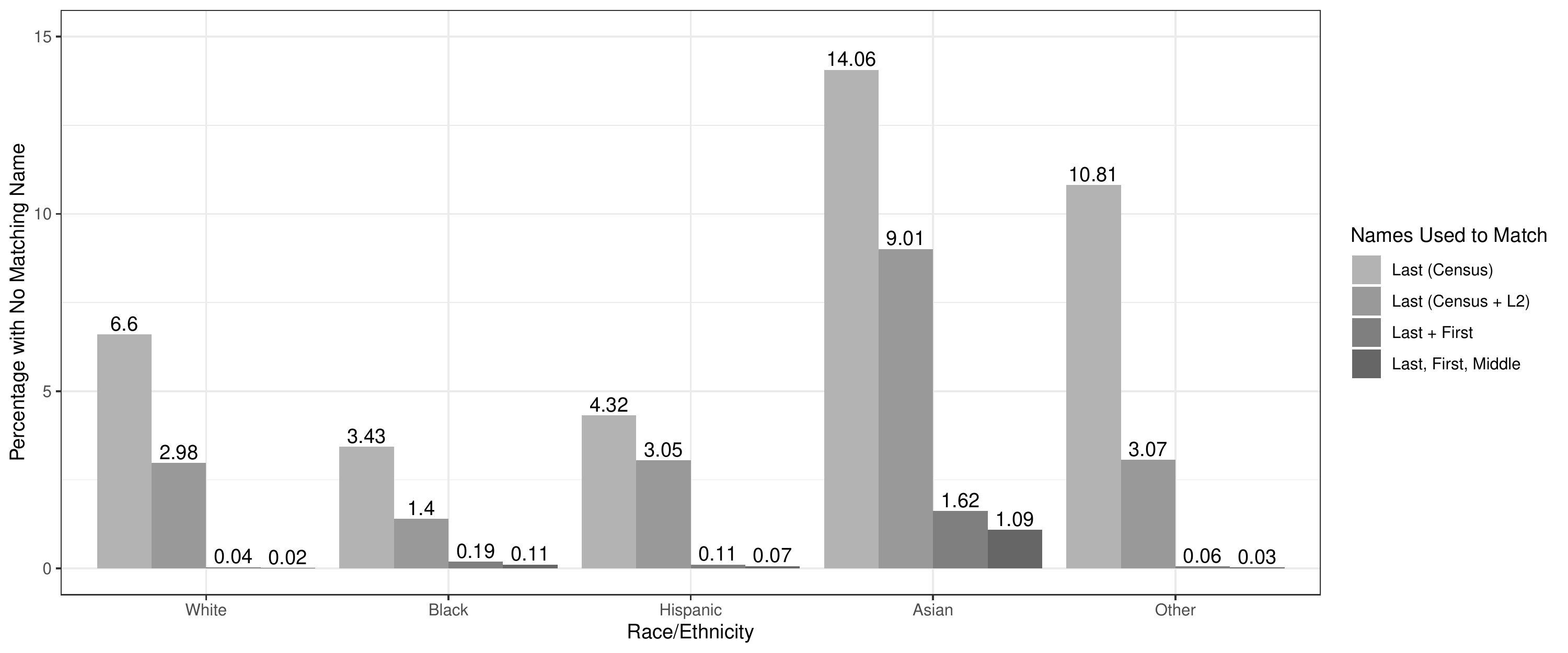}
  \caption{Percentage of individuals in each racial group who cannot be matched to \emph{any} name dictionary, under four different matching schemes: matching to census last names only; matching to census and L2 last names; matching last names and first names; and matching last, first, and middle names. Data is drawn from the voter files of Alabama, Florida, Georgia, Louisiana, North Carolina, and South Carolina.} 
  \label{fig:nameMatch}
\end{figure}

Because the voter file data is used both to compile the dictionaries
and to assess coverage, we iteratively hold out each of the six states
(Alabama, Florida, Georgia, Louisiana, North Carolina, and South
Carolina), and consider coverage using a dictionary compiled from the
other five states. The results in Figure \ref{fig:nameMatch} show that
dictionary augmentation --- and inclusion of additional names ---
substantially decreases the proportion of individuals who cannot be
matched to any dictionary. For non-Asian voters, all but a negligible
fraction of voters can be matched to at least one dictionary once all
their names are included. Among Asians, approximately one percent of
voters still cannot be matched when using first, middle, and last
names.  This, however, represents a dramatic improvement relative to
the case of exclusively using surnames, and sourcing data only from
the Census.

\section{Empirical Validation}
\label{sec:empirics}

To empirically validate our proposed improvements, we fit both the
standard BISG and our fBISG to the combined voter files from AL, FL,
GA, LA, NC, and SC, from L2, Inc.  As discussed in
Section~\ref{sec:problems}, this combined data set contains
information for roughly 38 million voters.

\subsection{The Setup}

In our validation, we treat the self-reported race of each
record as unobserved, and use the remaining available information for
that record to obtain posterior probability distributions over their
race.  Specifically, we use the last, first, and middle names of each
voter, as well as the Census block in which their reported home
address is located. We then compare predictions based on these
posterior distributions to the known racial categories of each record
in order to evaluate the overall quality of our fBISG predictions
\emph{vis-\`a-vis} those of the standard BISG methodology.

To obtain samples from the fBISG posterior distribution over races for
each voter in our combined voter file, we rely on the latest version
of the \texttt{wru} package in R \citep{wru}. We initialize the global
counts in the Gibbs updates of Equation~\eqref{eq:gibbs} using the
predictions based on the standard BISG methodology, and run a single
Markov chain for 1,500 iterations, discarding the first 500 samples as
burn-in.  Note that the conditional posterior in
Equation~\eqref{eq:gibbs} factorizes over locations $g$, which allows
us to fit the model separately across any level of geographic
aggregation defined on $\mathcal{G}$.

In our application, we fit models separately by state through the same
leave-one-out approach we used when assessing the impact of dictionary
coverage (see Section~\ref{subsec:names}). For instance, in sampling
the race probabilities of voters in North Carolina, we only use the
name-given-race distributions derived from augmenting the original
Census dictionary with records from all \emph{other} states.  This
ensures that the name-given-race distributions are not obtained from
the validation voter file. While the size of each voter file in our
sample did not require parallelization to make computation feasible,
factorization over $g$ allows researchers to parallelize their
analyses in order to fit our model efficiently on much larger data
files.  We completed all analyses on a laptop computer with an M1 Max
CPU and 64Gb of RAM in under 3 hours of wall time.

\subsection{Correcting the Zero Census Counts Problem}

\begin{figure}[t] \centering \spacingset{1}
  \includegraphics[width=\textwidth]{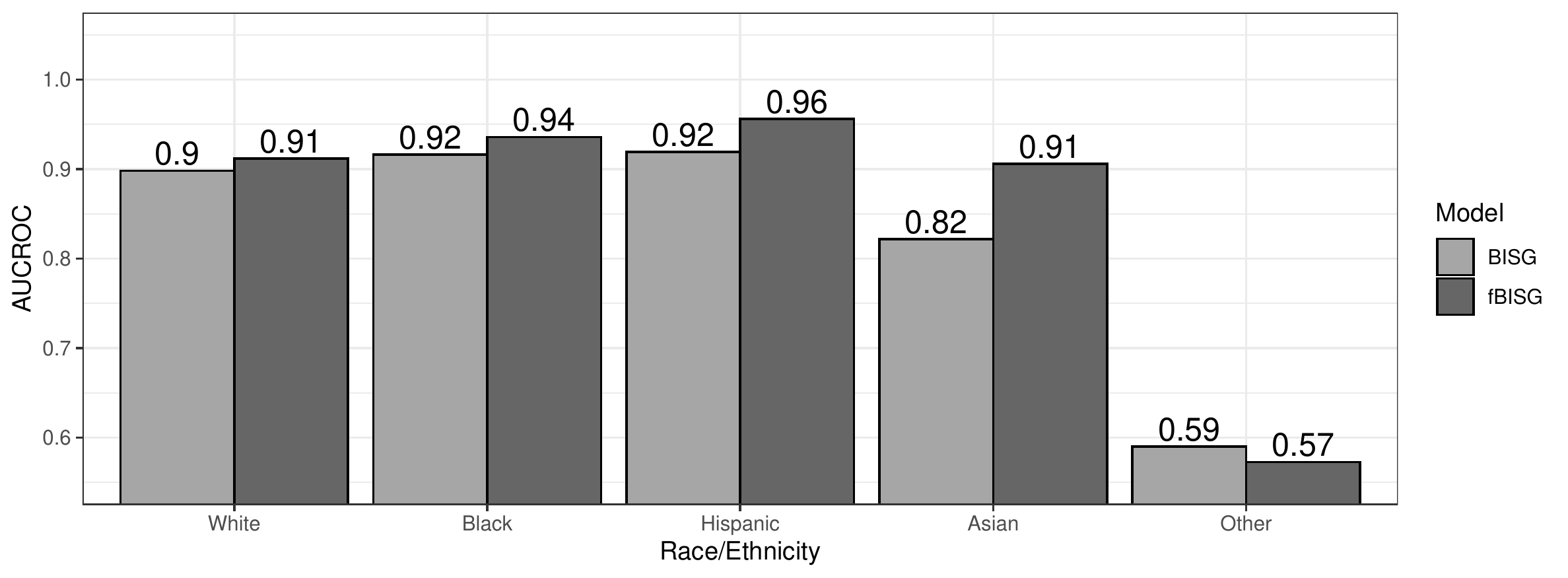}
  \caption{Area under the receiver operating characteristic curve for
    race predictions obtained using the standard BISG methodology
    (dotted) and our fully Bayesian BISG methodology (fBISG;
    stripped).  All results are based on the 2010 Census surname
    dictionary.  A greater value of AUROC indicates more accurate race
    classification.  For all but the ``Other'' category, the fBISG
    methodology has better classification performance than the
    standard BISG methodology, generating the most dramatic
    improvements among Asian minorities.}
  \label{fig:AUCROC-surnames}
\end{figure}

Figure~\ref{fig:AUCROC-surnames} shows, for each racial category, the
area under the receiver operating characteristic curve (AUROC) based
on posterior predictions generated by the standard BISG (light grey)
and fBISG (dark grey) methods.  For all but the ``Other'' racial
category, the predictive performance of the fBISG methodology
represents a substantial improvement over that of the standard
BISG. These performance gains are most dramatic for Hispanic and Asian
racial groups --- with the latter yielding an 11\% increase (from 0.82
using the BISG to 0.91 using the fBISG). In general, the use of fBISG
effectively eliminates the performance gap observed between major
racial categories when using BISG, which disproportionately affected
members of the Asian category.

The source of these improvements in classification accuracy varies by
racial category, as indicated by changes in False Positive and False
Negative error rates (see column ``Last name (census)'' of
Table~\ref{tab:ErrorRates} in the Appendix). Among White voters, fBISG
substantially reduces the false positive rate from 31\% to 24\%, while
keeping the false negative rate below 10\%. Among Black voters, the
improvement comes primarily from reducing false negatives, bringing
type II error down to about 24\% from the 36\% achieved by BISG.

In turn, while error reduction among Hispanics is small, accounting
for the zeroes in the Census counts substantially affects the accuracy
of classification among Asian voters. For the latter, fBISG reduces
both the false negative rate (from 47\% to 41\%) and the false
positive rate (from 0.78\% to 0.48\%) relative to the standard BISG
model. Given the large percentage (20\%) of Asian voters living in a
block for which the 2010 Census tallies register zero Asians, the
improvement induced by fBISG is unsurprising.

\begin{figure}[t]
\centering \spacingset{1}
\includegraphics[width=\textwidth, trim={1.2in 0 1.2in 0}, clip]{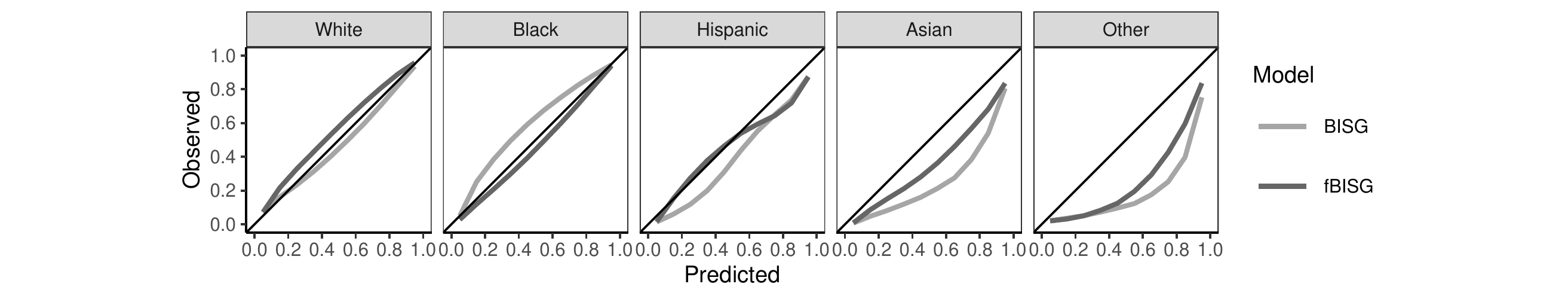}
\caption{Calibration curves for race predictions obtained using the
  standard BISG (dotted) and fBISG (dashed) methods.  The results are
  based on the 2010 Census surname dictionary. Curves closer to the
  45\textdegree line indicate better calibrated predictions.}
  \label{fig:Cal-Surnames}
\end{figure}

In addition to improving prediction accuracy, fBISG generally improves
the calibration of predicted probabilities, as shown in
Figure~\ref{fig:Cal-Surnames}. The figure shows predicted
probabilities vs. observed sample proportions of voters in each racial
category, for both BISG (light gray) and fBISG (dark gray)
methods. The closer the curves lie to the 45\textdegree line, the
better calibrated the corresponding method's predictions.  This is
because well-calibrated methods generate predicted probabilities that
match observed sample proportions of positive cases.  The figure shows
that fBISG can produce better calibrated predictions than the standard
BISG methodology for voters in most racial categories, producing well
calibrated predictions even among ``White'' voters --- the only
category in which calibration becomes slightly worse when making
predictions based on fBISG rather than on BISG. In sum, correcting the
zero-count measurement error issues can yield substantial improvements
in race predictive accuracy across all major racial categories.

\subsection{Correcting the Missing Race-Name Data Problem}

\begin{figure}[t!] \centering \spacingset{1}
\includegraphics[width=\textwidth]{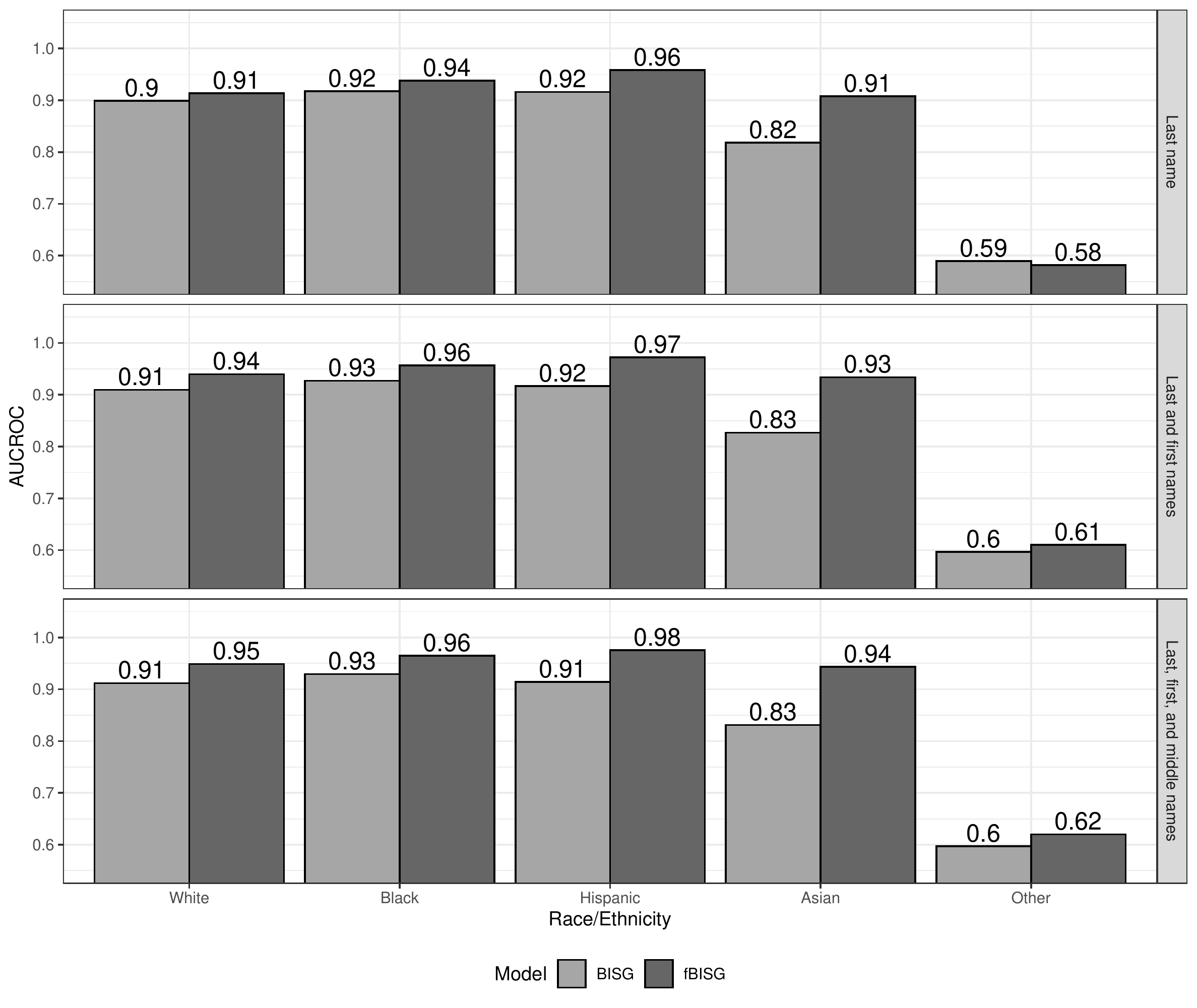}
\caption{Area under the receiver operating characteristic curve
  (AUROC) for race predictions obtained using the standard BISG (light
  grey) and fBISG (dark grey) methods.  The results are based on
  progressively more name information, starting with the L2-augmented
  surname dictionary (top panel). As before, higher values indicate
  better predictive accuracy. Overall, using more name information
  uniformly improves the accuracy of models, and using fBISG combined
  with more name information produces the most accurate models.}
  \label{fig:AUCROC-allnames}
\end{figure}

Correcting for name under-coverage, and adding additional name
information, also result in substantial improvements in predictive
accuracy. Figure~\ref{fig:AUCROC-allnames} shows the overall
improvement in predictive accuracy that results from using additional
name-race data from the L2 voter files, for both BISG (light bars) and
fBISG (dark bars). While these results aggregate across voter files,
recall that we mitigate over-fitting by sampling each state
separately, leaving names from that state out of the augmented
dictionaries.  Figure~\ref{fig:stateAUCROC-allnames} in the
Appendix presents the results separately for each state. 

Consistent with prior findings \citep[e.g.][]{voic:18}, we find that
using first and middle name information typically improves the
predictive performance of models. Moving from top to bottom, panels in
Figure~\ref{fig:AUCROC-allnames} show the steady improvement in model
performance as the predictions rely on an increasing amount of name
information (i.e., surnames only, surnames and first names, and
surnames, first names, and middle names together). While both BISG and
fBISG benefit from the progressively larger name-sets being used in
the prediction, fBISG is able to make the most of the additional
information. This is especially true among White voters, for whom
accuracy can be improved by as much as 4.4\% (from and AUROC of 0.91
to 0.95 using fBISG for generating predictions).

Once all our proposed solutions are implemented, improvements in
predictive accuracy over the standard BISG methodology are
substantial. Across major racial categories, the average increase in
AUROC is about 7\%, with improvements among Asian voters being as
large as 15\% (from 0.82 using the standard, surname-only BISG to 0.94
using all our proposed solutions). Using our fully specified model
renders predictive quality across major racial categories effectively
uniform, bringing probabilities of a correct classification over 0.95
for all major racial groups.

These gains in accuracy come primarily from substantial reductions in
the number of false negatives among all but White voters, as can be
seen by comparing the gray boxes within each error block across
columns 4 and 7 of Table~\ref{tab:ErrorRates} in the Appendix. For
non-White voters, type II error is reduced, on average, by 18
percentage points once all our solutions are implemented. These
improvements come primarily from correcting false positives attributed
to the White category, where we see a corresponding type I error rate
reduction of almost 16 percentage points.


\begin{figure}[t] \centering \spacingset{1}
\includegraphics[width=\textwidth, trim={1.2in 0 1.2in 0in}, clip]{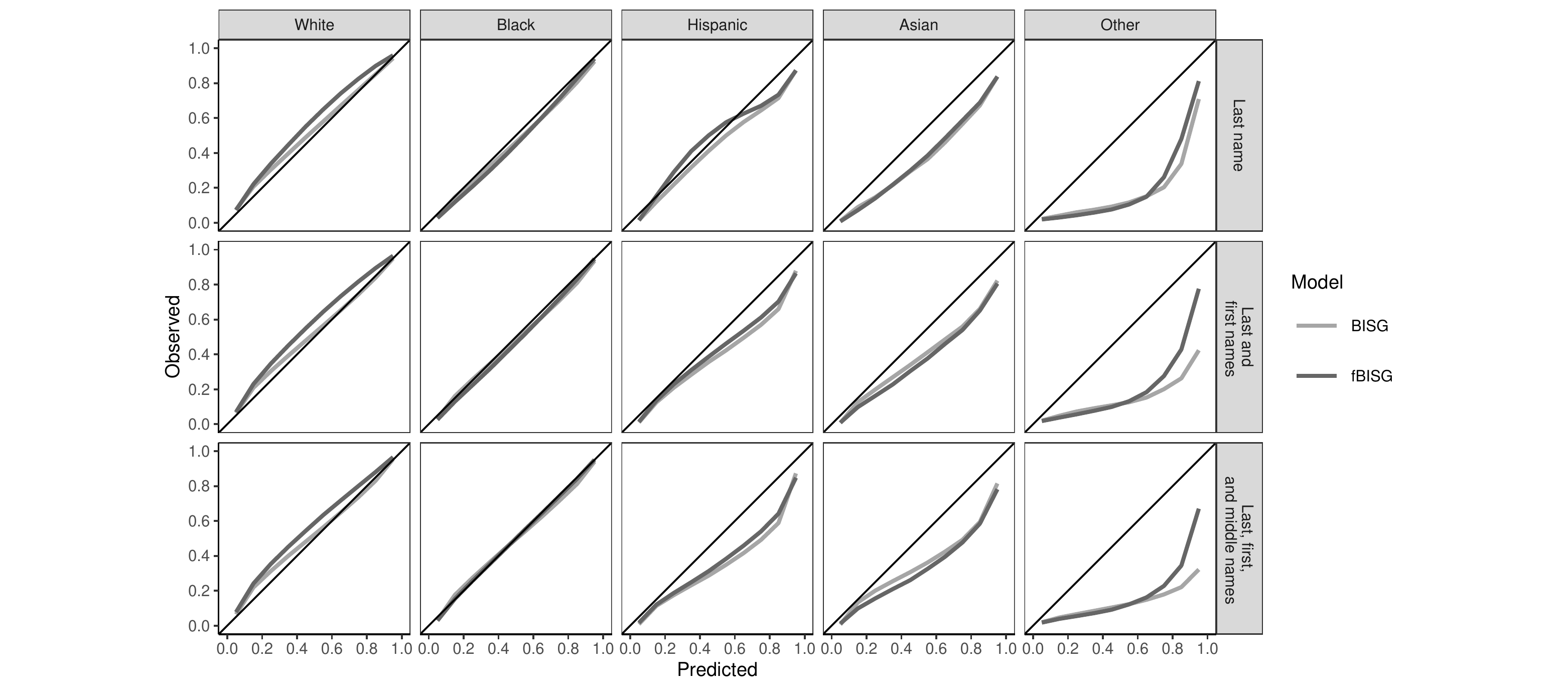}
\caption{Calibration curves for race predictions obtained using BISG
  (light grey) and fBISG (dark grey), using progressively more name
  information from dictionaries augmented with L2 data (from top to
  bottom rows). As before, curves closer to the 45\textdegree\ line
  indicate better calibrated predictions.  }
  \label{fig:Cal-AllNames}
\end{figure}

Augmenting name dictionaries can also improve calibration, although
gains are more modest on this front.  Figure~\ref{fig:Cal-AllNames}
presents the calibration curves for the BISG (light gray) and fBISG
(dark gray) models, now estimated using augmented name
dictionaries. Comparing the first row of this figure to the panels in
Figure~\ref{fig:Cal-Surnames}, we find that using the expanded surname
set can improve name calibration --- particularly among White and
Black voters, for whom calibration is exceptionally high --- and
reduce the observed differences between BISG and fBISG approaches.

Moreover, gains in accuracy from adding information on first and
middle names are not made at the expense of calibration, with
calibration curves that are effectively the same across most rows of
Figure~\ref{fig:Cal-AllNames}. The exceptions to this pattern come
from the inclusion of middle names among Hispanic and Asian voters,
which slightly worsens calibration for predictions based on both BISG
and fBISG. This is likely the result of different norms around middle
name usage among members of these racial groups. Of all four major
racial categories, calibration remains worst among Asian voters.

\section{Conclusion}

In this paper, we consider the problem of predicting an individual's
race. This task is especially relevant to modern research on racial
equity in areas such as public health and political science. The
current state-of-the-art approach is Bayesian Improved Surname
Geocoding (BISG), which uses surname and geolocation data to generate
a probabilistic prediction for each individual over racial
classes. Yet, as we have shown, BISG predictions can underperform for
minority groups due to two consistent challenges: inaccurate Census
counts, and name under-coverage.

To address these challenges, we have introduced a fully Bayesian
analogue known as fully-Bayesian Improved Surname Geocoding (fBISG)
that addresses the problem of Census zero counts. Moreover, we have
augmented our name dictionaries, including additional surnames, as
well as first and middle names, sourced from voter files in six
southern states provided by L2, Inc. Taken together, these
methodological improvements yield substantial performance gains in
predictive accuracy --- as measured by AUROC, as well as false positive
and false negative error rates under maximum a posteriori predictions
-- while simultaneously improving the calibration of
predictions. Moreover, the gains are most pronounced among Hispanics
and Asian Americans, drawing their predictions almost to parity with
those for White and Black voters in terms of accuracy. We believe
these improvements will be useful for practitioners, allowing them to
obtain improved individual-level racial predictions and better
characterize disparate racial impacts.

\bigskip
\pdfbookmark[1]{References}{References} \spacingset{1.5}
\bibliography{my,imai,biblio}

\newpage
\appendix

\setcounter{equation}{0}
\setcounter{figure}{0}
\setcounter{table}{0}
\setcounter{section}{0}
\renewcommand {\theequation} {A\arabic{equation}}
\renewcommand {\thefigure} {A\arabic{figure}}
\renewcommand {\thetable} {A\arabic{table}}

\section*{Appendix: Additional Results}

\begin{table}[h]
\centering\spacingset{1}
\begin{tabular}{llrrr}
\toprule
\textbf{Ethnicity} & \textbf{Data}  & \textbf{\renewcommand{\arraystretch}{0.75} \begin{tabular}[c]{@{}c@{}}Nonzero Census\\ Blocks\end{tabular}} & \textbf{\renewcommand{\arraystretch}{0.75} \begin{tabular}[c]{@{}c@{}}Zero Census\\ Blocks\end{tabular}} & \multicolumn{1}{l}{\textbf{Total}} \\ \midrule
Overall Error Rate \hspace{5mm} &                & 14.5\%    & 100\%     & 16.9\%  \\ \hline
\multirow{2}{*}{White}              & False negative & 5.6\%     & 100\%   & 6.1\%   \\
             & False positive & 31.4\%    & NA  & 31.4\%  \\ \hline
\multirow{2}{*}{Black}              & False negative & 33.7\%    & 100\%  & 36.4\%  \\
            & False positive & 3.5\%    & NA  & 3.5\%  \\  \hline
\multirow{2}{*}{Hispanic}           & False negative & 15.7\%    & 100\%  & 20.3\%  \\
         & False positive & 2.2\%    & NA & 2.2\%  \\ \hline
\multirow{2}{*}{Asian}              & False negative & 33.2\%    & 100\%  & 46.6\%  \\
             & False positive & 0.7\%    & NA  & 0.7\%  \\ \hline
\multirow{2}{*}{Other}              & False negative & 92.7\%    & 100\%  & 94.9\%  \\
              & False positive & 0.3\%    & NA  & 0.3\%  \\ \bottomrule
\end{tabular}
\bigskip
\caption{Overall classification error rate as well as false positive
  (Type I error) and false negative (Type II error) rates for White,
  Black, Latino, Asian, and Other voters using the standard BISG
  prediction as implemented in the \texttt{WRU} package. Each voter is
  classified to the racial category with the highest predicted
  probability. We compare rates for individuals in blocks for whom the
  Census sets a nonzero prior for their true racial group (``Nonzero
  Census Blocks'') against individuals in blocks from who the census
  sets a zero prior (``Zero Census Blocks''). All individuals are
  classified to the wrong racial group in zero Census blocks, so false
  negative rates are 100\% while false positive rates are undefined.} 
\label{tab:censusBlockPreds}
\end{table}

\begin{table}[h]
\centering\spacingset{1}
\begin{tabular}{llrrr}
\toprule
\textbf{Ethnicity} & \textbf{Error}  & \textbf{\renewcommand{\arraystretch}{0.75} \begin{tabular}[c]{@{}c@{}}Name Matched\\ to Dictionary\end{tabular}} & \textbf{\renewcommand{\arraystretch}{0.75} \begin{tabular}[c]{@{}c@{}}Name Unmatched\\ to Dictionary\end{tabular}} & \multicolumn{1}{l}{\textbf{Total}} \\ \midrule
Overall Error Rate \hspace{5mm} &                & 16.4\%    & 24.7\%  & 16.9\%  \\ \hline
\multirow{2}{*}{White}              & False negative & 6.1\%     & 6.5\%   & 6.1\%   \\
            & False positive & 30.1\%    & 58.8\%  & 31.4\%  \\ \hline
\multirow{2}{*}{Black}              & False negative & 35.5\%    & 63.7\%  & 36.4\%  \\
            & False positive & 3.6\%    & 2.1\%  & 3.5\%  \\ \hline
\multirow{2}{*}{Hispanic}           & False negative & 18.4\%    & 65.4\%  & 20.3\%  \\
           & False positive & 2.1\%    & 4.1\%  & 2.2\%  \\ \hline
\multirow{2}{*}{Asian}              & False negative & 40.3\%    & 84.6\%  & 46.6\%  \\
              & False positive & 0.6\%    & 2.7\%  & 0.7\%  \\ \hline
\multirow{2}{*}{Other}              & False negative & 94.4\%    & 99.3\%  & 94.9\%  \\
              & False positive & 0.3\%    & 0.2\%  & 0.3\%  \\ \bottomrule
\end{tabular}
\bigskip
\caption{Overall classification error rate as well as false positive
  (Type I error) and false negative (Type II error) rates for White,
  Black, Latino, Asian, and Other voters using prediction using
  standard BISG as implemented in the \texttt{WRU} package. Each voter
  is classified to the racial category with the highest predicted
  probability. We compare rates for individuals whose names are
  matched to our name dictionary against those whose names are not
  matched (in which case a national racial prior is used). Error rates
  are significantly higher among those whose names are unmatched.} 
\label{tab:missingNamePreds}
\end{table}

\begin{table}[ht]
\centering\spacingset{1}
\begin{tabular}{lllcccc}
\toprule
\textbf{Ethnicity} & \textbf{Error} & \textbf{Model} & \textbf{\renewcommand{\arraystretch}{0.75} \begin{tabular}[c]{@{}c@{}}Last name \\(Census) \end{tabular}} &  \textbf{\renewcommand{\arraystretch}{0.75} \begin{tabular}[c]{@{}c@{}}Last name \\ (augmented)\end{tabular}} & \textbf{\renewcommand{\arraystretch}{0.75} \begin{tabular}[c]{@{}c@{}} Last \& first\\ names  \end{tabular}} & \textbf{\renewcommand{\arraystretch}{0.75} \begin{tabular}[c]{@{}c@{}} Last, first, \&\\middle names \end{tabular}} \\
\midrule
\multirow{ 4}{*}{White}    & \multirow{2}{*}{False negative} & BISG  & {\cellcolor{gray!25}6.11}  & 8.60  & 6.87  & 6.29  \\
    &  & fBISG & 9.37  & 9.67  & 7.79  & {\cellcolor{gray!25}6.93} \\ \cline{3-7}
    & \multirow{2}{*}{False positive} & BISG  & {\cellcolor{gray!25}31.40} & 25.36 & 21.90 & 19.73 \\
    &  & fBISG & 24.25 & 23.00 & 17.69 & {\cellcolor{gray!25}15.59} \\
\midrule
\multirow{ 4}{*}{Black}     & \multirow{2}{*}{False negative} & BISG  & {\cellcolor{gray!25}36.44} & 26.22 & 22.98 & 21.24 \\
    &  & fBISG & 24.39 & 23.81 & 19.21 & {\cellcolor{gray!25}17.93} \\ \cline{3-7}
    &  \multirow{2}{*}{False positive} & BISG  & {\cellcolor{gray!25}3.54}  & 6.29  & 4.81  & 4.19  \\
    &  & fBISG & 6.91  & 6.89  & 5.25  & {\cellcolor{gray!25}4.23}  \\
\midrule  
\multirow{ 4}{*}{Hispanic}  & \multirow{2}{*}{False negative} & BISG  & {\cellcolor{gray!25}20.35} & 24.58 & 20.21 & 18.38 \\
 & & fBISG & 24.16 & 22.81 & 14.99 & {\cellcolor{gray!25}12.05} \\ \cline{3-7}
 &  \multirow{2}{*}{False positive} & BISG  & {\cellcolor{gray!25}2.22}  & 2.05  & 1.92  & 1.99  \\
 &  & fBISG & 2.11  & 2.08  & 2.06  & {\cellcolor{gray!25}2.19}  \\
\midrule
\multirow{ 4}{*}{Asian}    & \multirow{2}{*}{False negative} & BISG  & {\cellcolor{gray!25}46.59} & 48.36 & 45.27 & 43.68 \\
    &  & fBISG & 41.10 & 39.35 & 32.58 & {\cellcolor{gray!25}29.40} \\ \cline{3-7}
    &  \multirow{2}{*}{False positive} & BISG  & {\cellcolor{gray!25}0.74}  & 0.41  & 0.39  & 0.40  \\
    & & fBISG & 0.48  & 0.52  & 0.54  & {\cellcolor{gray!25}0.60}  \\
\midrule
\multirow{ 4}{*}{Other}    & \multirow{2}{*}{False negative} & BISG  & {\cellcolor{gray!25}94.90} & 94.30 & 92.98 & 91.96 \\
    & & fBISG & 95.21 & 94.10 & 92.91 & {\cellcolor{gray!25}91.85} \\ \cline{3-7}
    & \multirow{2}{*}{False positive} & BISG  & {\cellcolor{gray!25}0.28}  & 0.43  & 0.63  & 0.83  \\
    &  & fBISG & 0.16  & 0.50  & 0.55  & {\cellcolor{gray!25}0.73}  \\  
\bottomrule
\end{tabular}
\caption{False positive (Type I error) and false negative (Type II
  error) error rates for White, Black, Latino, Asian, and Other voters
  using predictions from standard BISG and from our proposed fBISG
  model. Each voter is classified to the racial category with the
  highest posterior probability. For all but the `Other' category,
  both types of errors are reduced as you move from standard,
  Census-dictionary BISG, to fBISG using the augmented dictionary
  (gray cells within each block of rows).} 
\label{tab:ErrorRates}
\end{table}

\begin{figure}[h]
  \centering\spacingset{1}
  \includegraphics[width=\textwidth]{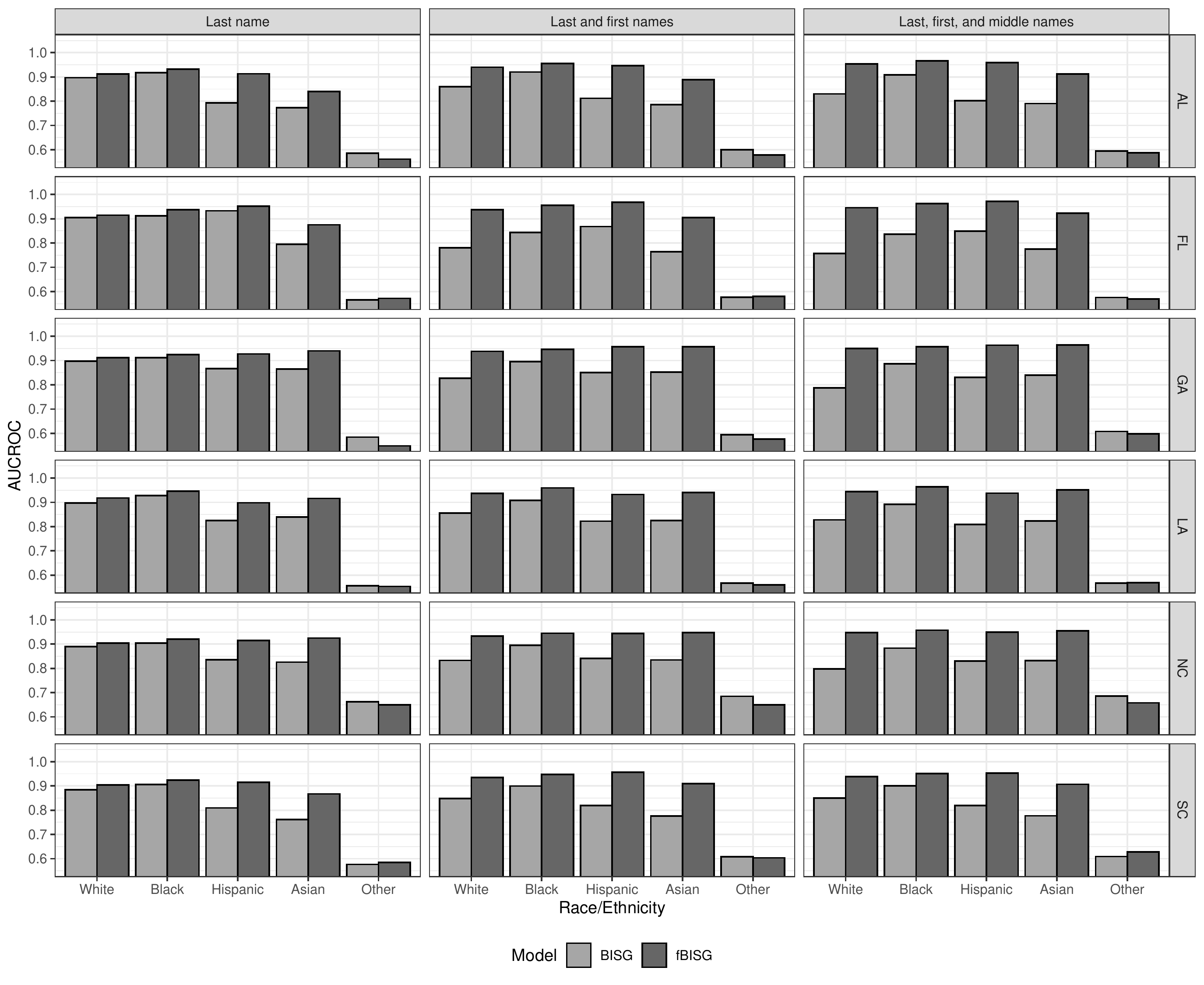}
  \caption{Area under the receiver operating characteristic curve
  (AUROC) for race predictions obtained using the standard BISG (light
  grey) and fBISG (dark grey) methods, by state.  The results are based on
  progressively more name information, starting with the L2-augmented
  surname dictionary (left-most panel). Higher values indicate
  better predictive accuracy. Overall, the same patterns we observed when considering
  all states combined are evident when we disaggregate accuracy by state: additional name
  information improves accuracy, especially when using fBISG.}
  \label{fig:stateAUCROC-allnames}
\end{figure}

\end{document}